\documentclass[10pt,twocolumn,letterpaper]{article}

\usepackage{wacv}              

\usepackage{graphicx}
\usepackage{amsmath}
\usepackage{amssymb}
\usepackage{booktabs}
\usepackage{bm}
\usepackage{graphics} 
\usepackage{epsfig} 
\usepackage{hyperref}
\usepackage{balance}
\usepackage{mathptmx} 
\usepackage{times} 
\usepackage{interval}
\usepackage{threeparttable}
\usepackage{adjustbox}
\usepackage{multicol}
\usepackage{lipsum}
\usepackage{caption}
\usepackage{pgfplotstable}
\usepackage{subcaption}
\usepackage{tikz}
\usepackage{array}
\usepackage{pgfplots}
\usepackage{multirow}

\usepackage[capitalize]{cleveref}
\crefname{section}{Sec.}{Secs.}
\Crefname{section}{Section}{Sections}
\Crefname{table}{Table}{Tables}
\crefname{table}{Tab.}{Tabs.}

\def\wacvPaperID{286} 
\def\confName{WACV}
\def\confYear{2025}

\begin{document}

\title{PACA: \underline{P}erspective-\underline{A}ware \underline{C}ross-\underline{A}ttention Representation \\ for Zero-shot Scene Rearrangement}

\author{
Shutong Jin$^{1*}$, Ruiyu Wang$^{1*}$, Kuangyi Chen$^2$, Florian T. Pokorny$^1$ \\
$^1$KTH Royal Institute of Technology, \ $^2$Graz University of Technology\\
{\tt\small \{shutong, ruiyuw, fpokorny\}@kth.se, kuangyi.chen@tugraz.at}\\
}
\maketitle

\begingroup
\renewcommand\thefootnote{} 
\footnotetext{*Equal contribution; This work was partially supported by the Wallenberg AI, Autonomous Systems and Software Program (WASP) funded by the Knut and Alice Wallenberg Foundation. The computations were enabled by the supercomputing resource Berzelius provided by National Supercomputer Centre at Linköping University and the Knut and Alice Wallenberg Foundation, Sweden.}
\addtocounter{footnote}{-1} 
\endgroup

\begin{abstract}
   Scene rearrangement, like table tidying, is a challenging task in robotic manipulation due to the complexity of predicting diverse object arrangements. Web-scale trained generative models such as Stable Diffusion~\cite{rombach2022high} can aid by generating natural scenes as goals. To facilitate robot execution, object-level representations must be extracted to match the real scenes with the generated goals and to calculate object pose transformations. Current methods typically use a multi-step design that involves separate models for generation, segmentation, and feature encoding, which can lead to a low success rate due to error accumulation. Furthermore, they lack control over the viewing perspectives of the generated goals, restricting the tasks to 3-DoF settings. In this paper, we propose \textit{PACA}, a zero-shot pipeline for scene rearrangement that leverages perspective-aware cross-attention representation derived from Stable Diffusion. Specifically, we develop an object-level representation that integrates generation, segmentation, and feature encoding into a single step. Additionally, we introduce perspective control, thus enabling the matching of 6-DoF camera views and extending past approaches that were limited to 3-DoF top-down settings. The efficacy of our method is demonstrated through its zero-shot performance in real robot experiments across various scenes, achieving an average matching accuracy and execution success rate of 87\% and 67\%, respectively. 
\end{abstract}

\section{Introduction}
\label{sec:intro}
\begin{figure}[!t]
\centering
\includegraphics[height=7.2cm]{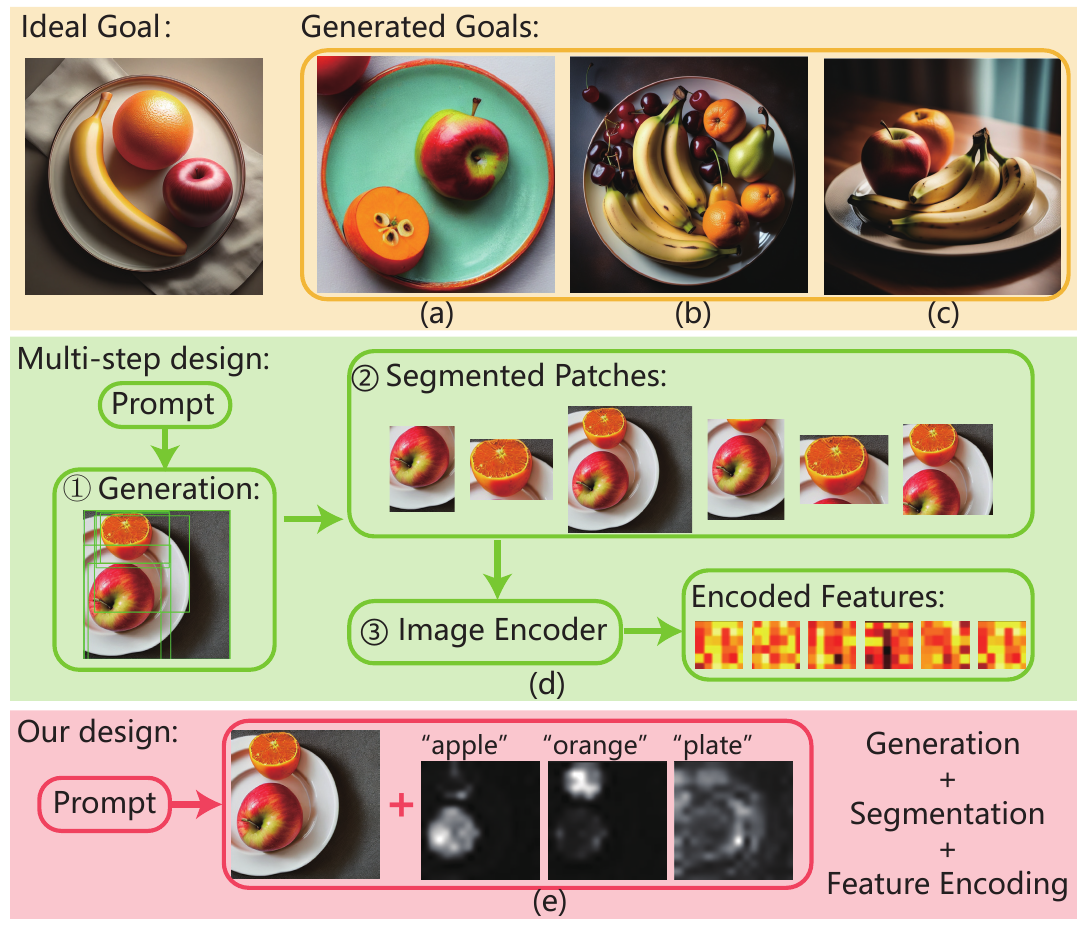}
\caption{The images were generated with Stable Diffusion and the prompt ``One apple, one orange, and one banana in the plate''. The generated images exhibit several issues: (a) distortion, (b) introduction of new objects, and (c) unmatched image perspectives.   An illustration of multi-step design: (d) Generation, segmentation, and feature encoding are performed using three different models. During segmentation, repetitive cropping occurs due to the distortion in generated goals, introducing errors to the subsequent feature encoding process. The segmented patches were produced by Mask R-CNN~\cite{he2017mask}. Our single-step design: (e) Segmentation and feature encoding are conducted concurrently with the generation.}
\label{fig:first_figure}
\end{figure}

Scene rearrangement is a highly modular task in robotic manipulation and typically comprises several phases: representing the scene, predicting goals using the chosen representations, matching the representations between the scene and the goals, and determining object pose transformations for robot execution~\cite{batra2020rearrangement}. The first challenge lies in predicting goals, as objects in the scene can vary widely and their ideal spatial relationships may differ significantly based on the context~\cite{paxton2022predicting}. For example, when organizing cutlery, an ideal arrangement might consist of a plate in the middle with a knife and fork on each side of the plate. However, when shifting to tidying fruits, the preferred arrangement could be placing all the fruits on a plate. This variability significantly increases the cost of collecting large-scale real-world training datasets with example arrangements of poses favored by humans. Consequently, current efforts are focused on generating datasets within simulated environments that have a limited number of predefined patterns~\cite{liu2022structformer, jain2023transformers, liu2022structdiffusion}. However, this approach may face challenges when encountering unseen objects or environments.


Conversely, the text-driven generative models like Stable Diffusion~\cite{rombach2022high} and DALL·E 3~\cite{betker2023improving} have the potential to address the aforementioned challenge of goal prediction with their web-scale training datasets of scenes from typical daily scenarios. To match objects in the real scene with the generated goal and to calculate pose transformations, object-level representations need to be derived through separate generation, segmentation and feature encoding~\cite{kapelyukh2023dall, kapelyukh2023dream2real}. However, due to the stochastic sampling~\cite{ho2020denoising}, generated images often exhibit distortions~\cite{zhang2023text} (\textit{Figure}~\ref{fig:first_figure} (a)). Furthermore, the training data for the generative models usually depict objects as groups within a scene rather than as isolated items~\cite{schuhmann2021laion}. This can lead to the unintended occurrences of objects not specified in the user's prompt during goal generation (\textit{Figure}~\ref{fig:first_figure} (b)). Both factors contribute to degraded segmentations of the generated image, as reflected in repetitive and inaccurate cropping (\textit{Figure}~\ref{fig:first_figure} (d)). Feature encoding with these degraded segmentations results in error accumulation during matching. And the introduction of unintended objects exacerbates this situation by causing excessive matching. For simplicity, we refer to the above issues as multi-step error accumulation.


Additionally, the lack of control over the generated image perspective makes matching with the real scene extremely challenging (\textit{Figure}~\ref{fig:first_figure} (c)), which has led past image-goal-based works~\cite{gkanatsios2023energy, kapelyukh2023dall} to focus on the comparatively simpler top-down 3-DoF settings. In this paper, we aim to address the multi-step error accumulation and the 3-DoF limitation to maximize the potential of web-scale trained generative models. Our contributions are:
\begin{itemize}
    \item We introduce \textit{PACA}, a training-free pipeline for scene rearrangement that utilizes web-scale trained Stable Diffusion~\cite{rombach2022high}. 
    \item Leveraging the lossy denoising process of diffusion models, we develop an object-level representation that utilizes cross-attention at different denoising stages, integrating generation, segmentation, and feature encoding into a single step (\textit{Figure}~\ref{fig:first_figure} (e)). 
    \item By introducing additional perspective control during the generation, we expand the image-goal-based method from 3-DoF to 6-DoF.
    \item We demonstrate the efficacy of \textit{PACA} through its zero-shot performance in real robot experiments across various scenes.
\end{itemize}


\section{Related Work}
\subsection{Text-driven Generative Models for Robotic \mbox{Manipulation}}

Robotic manipulation and scene rearrangement leveraging text-driven generative models share a common framework of generating visual goals and deriving robot actions from these visual goals~\cite{black2023zero, gao2024can, chen2023genaug, yu2023scaling, mandi2022cacti}. The key distinction is that for robotic manipulation, the robot arm and gripper must be visible in the generated visual goals, which are typically sequences of images~\cite{ko2023learning, du2024learning, bharadhwaj2023zero, ajay2024compositional}. Robot actions are then derived from these visual goals by training inverse dynamics models. Prominent examples include \textit{Black et al.}~\cite{black2023zero} who finetune an image-editing diffusion model~\cite{brooks2023instructpix2pix} on video data to produce image subgoals and train a reaching policy to carry out the goals in a stepwise manner; \textit{Ko et al.}~\cite{ko2023learning} synthesize a full robot execution video and predict the robot actions from the computed optical flow~\cite{xu2022gmflow} between frames. 

Conversely, scene rearrangement only requires the presence of the objects to be rearranged in the visual goal, which is typically a single image generated by DALL·E\cite{betker2023improving} or Stable Diffusion~\cite{rombach2022high}. While this approach eliminates the need for costly large-scale training with human demonstrations~\cite{ebert2021bridge, robomimic2021}, it requires detailed object-level representations computed from the goal image for action derivation. 

\subsection{Goal Prediction in Scene Rearrangement}

Scene rearrangement can be tackled with different focuses (e.g. visual perception~
\cite{migimatsu2022grounding}, motion planning~\cite{curtis2022long}), with goal prediction consistently serving as a core element. Various methods for predicting goals have been investigated, including energy models~\cite{gkanatsios2023energy, du2020compositional}, scene graph generations~\cite{zhai2023sg, kulshrestha2023structural}, object-object relations~\cite{paxton2022predicting, yuan2022sornet}, gradient field~\cite{wu2022targf}, etc. Most of these methods rely on extensive training on large-scale datasets and generate outputs based on the learned distributions~\cite {wei2023lego}. The cost of dataset collection limits these methods to simulation, where goals are generated by random object swapping~\cite {qureshi2021nerp} or following fixed patterns~\cite{liu2022structformer, liu2022structdiffusion}. However, this can lead to a sim-to-real gap~\cite{hofer2021sim2real} and generalization issues when encountering unseen objects in reality. To tackle this problem, we adopt generative models~\cite{rombach2022high, betker2023improving, saharia2022photorealistic} trained on web-scale data to generate natural scenes as the goal.


\textbf{Zero-shot Scene Rearrangement.}
Compared to large-scale training methods, zero-shot approaches~\cite{kapelyukh2023dall, gkanatsios2023energy, newman2024leveraging, kapelyukh2023dream2real} are still limited. Prominent examples include DALL·E-Bot~\cite{kapelyukh2023dall}, which segments real scenes into object masks~\cite{he2017mask}, encodes their semantics using CLIP~\cite{radford2021learning} and then captions them to generate goal images with DALL·E 2~\cite{ramesh2022hierarchical}; SREM~\cite{gkanatsios2023energy} generates compositional configurations by minimizing energy functions constructed through a semantic parser~\cite{dong2016language, brown2020language}. However, those approaches are confined to 3-DoF top-down settings due to the difficulty of matching different image perspectives. Dream2Real~\cite{kapelyukh2023dream2real} achieves 6-DoF rearrangement by rendering scenes in 3D using NeRFs~\cite{mildenhall2021nerf} and utilizing other foundational models~\cite{kirillov2023segment, openai2023gpt, cheng2022xmem, curless1996volumetric, radford2021learning}. Nonetheless, the multi-step design of these methods introduces accumulated errors in real-world scenarios when one of the components performs sub-optimally. In this paper, we propose to extend existing methods from 3 DoF to 6 DoF by imposing perspective control during the image goal generation, with integrated segmentation and feature encoding. 
\begin{figure*}[!t]
\centering
\includegraphics[height=4.5cm]{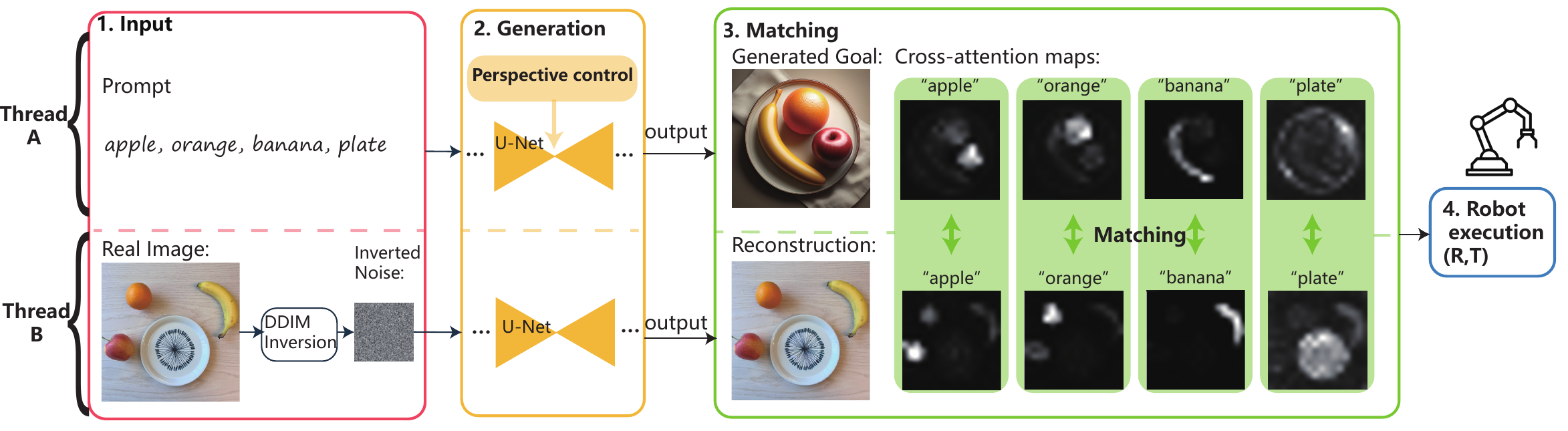}
\vspace{-0.5em}
\caption{Pipeline of the proposed \textit{PACA}. Generation, segmentation and feature encoding are integrated into a single step. The prompt is directly utilized to produce the generated goal. The real image is first inverted to its noisy counterpart and then reconstructed to extract segmentations and features for matching. The cross-attention maps highlight the segmentation and encoded features specific to each object's descriptors, thereby facilitating the matching process and transformation calculations.}
\label{fig:structure}
\vspace{-1em}
\end{figure*}

\section{Preliminary}
\subsection{\textbf{Diffusion models}} Denoising Diffusion Probabilistic Models (DDPMs) \cite{ho2020denoising} use a U-Net \cite{ronneberger2015u} architecture, denoted by $\epsilon_{\theta}$, to model a data distribution $x \sim q(x_0)$. The \textit{forward process} is fixed to a Markov chain where Gaussian noise is gradually added to the data $x$:
\begin{equation}
    q(x_t|x_{t-1}) = \mathcal{N}(x_t; \sqrt{1-\beta_t}x_{t-1},\; \beta_t \mathbf{I}),
    \label{eq:ddpm_diffusing}
\end{equation}
where $\beta_t$ is a variance scheduler learned by reparameterization~\cite{kingma2013auto}. Due to the property of the Markov chain, sampling $x_t$ at an arbitrary timestep $t$ can be obtained by:
\begin{equation}
    q(x_t|x_0) = \mathcal{N}(x_t; \sqrt{\bar{\alpha}_t}x_0,\;(1 -  \bar{\alpha}_t)\mathbf{I}),
    \label{eq:ddpm_arbitrary}
\end{equation}
where $\alpha_t = 1 - \beta_t$ and $\bar{\alpha}_t = \prod_{s=1}^{t} \alpha_s$. 
In the \textit{reverse process} or \textit{denoising process}, $x_0 \sim p_{\theta}(x_0)$ is sampled from the learned Gaussian transition:
\begin{equation}
    p_{\theta}(x_{t-1} | x_t) \sim \mathcal{N}(x_{t-1}; \mu_{\theta}(x_t, t),\; \Sigma_{\theta}(x_t,\;t)),
    \label{ddpm_reverse}
\end{equation}
where $\Sigma_{\theta}(x_t,\;t)$ are untrained time dependent constants and $\mu_{\theta}(x_t, \;t)$ can be computed from $\varepsilon_\theta (x_t,\; t)$. The training objective can be derived as minimizing:
\begin{equation}
\mathcal{L}(\theta) = \mathbb{E}_{x_0\sim q(x_0),\; \epsilon\sim\mathcal{N}(0,\;\mathbf{I}),\;t}\left[ \left\| \epsilon - \epsilon_{\theta}(x_t, \;t) \right\|^2 \right].
\label{eq:ddpm}
\end{equation}


\subsection{\textbf{Text-driven Latent Diffusion Models}} Latent Diffusion Models (LDMs) \cite{rombach2022high} are variants of DDPMs where the \textit{forward process} and \textit{denoising process} operate in the latent space. The training objective of text-driven LDMs is:

\begin{equation}
    \mathcal{L}_{t}(\theta) = \mathbb{E}_{z_0\sim q(\mathcal{E}(x_{0})), \;\epsilon\sim\mathcal{N}(0,\;\mathbf{I}),\;t}\left[ \left\|\epsilon - \varepsilon_\theta (z_t,\; t,\; p)\right\|^{2}\right],
    \label{eq:text-driven ldm}
\end{equation}
where $ z = \mathcal{E}(x) $, with encoder $ \mathcal{E} $, can later be mapped back to pixel space by a decoder $ \mathcal{D} $. $ P $ is the text input that serves as guidance, encoded as $ p = \psi(P) $.
 
\subsection{\textbf{Adding additional control}} Additional control $C \in \mathbb{R}^{h \times w \times c}$, with $\{h, w, c\}$ as the height, width, and number of channels, can be added by connecting a trainable copy to the frozen LDMs with zero convolution layers~\cite{zhang2023adding}. The training loss is: 
\begin{equation}
    \mathcal{L}_{c}(\theta) = \mathbb{E}_{z_0\sim q(\mathcal{E}(x_{0})), \;\epsilon\sim\mathcal{N}(0,\;\mathbf{I}),\;t} \left[ \left\lVert \epsilon - \varepsilon_\theta(z_t,\; t,\; p,\; C) \right\rVert^2 \right].
    \label{eq:control_ldm}
\end{equation}

\section{Method}
\subsection{Problem Description}
The scene rearrangement task involves transitioning an initial scene \( X \) to a human-favored scene \( X^* \), by transforming the existing objects through an embodied agent, such as a tabletop robotic arm.


\textit{Figure}~\ref{fig:structure} illustrates the pipeline of our proposed \textit{PACA}, which structures the scene rearrangement task into four main steps:  
1) \textbf{Input acquisition}, consisting of two threads: a) the user's prompt $P$; b) an RGBD image $X^{real}$ of the current real scene. 
2) \textbf{Goal generation} with perspective control, while simultaneously generating object-level representations. 
3) \textbf{Matching} the generated representations between the goal and the real scene to determine the pose transformation for each object.
4) \textbf{Robot execution} with calculated object pose transformation.
In the following sections, we will delve into the details of these steps.

\subsection{Generation with prompt $P$: Thread A}

We denote the user prompt that includes object categories as $ P $, along with a random seed $ s $ that can be used to reproduce the generated scene if needed. The prompt $ P $ is fed into Stable Diffusion~\cite{rombach2022high}, following the training objective in \textit{Eq.}~\ref{eq:text-driven ldm}. The generated image is denoted as $ x^{goal} $. 
\begin{equation}
    \text{3-DoF:\quad} x^{goal} = \textsc{StableDiffusion}(P, \;s, \; \beta_{cfg}),
    \label{eq: stablediffusion}
\end{equation}
where 3-DoF refers to the top-down view generation, and $ \beta_{cfg} $ is the Classifier-Free Guidance (CFG) scale, fixed at 7.5 during generation.

\begin{figure}[h]
\centering
\includegraphics[height=2.5cm]{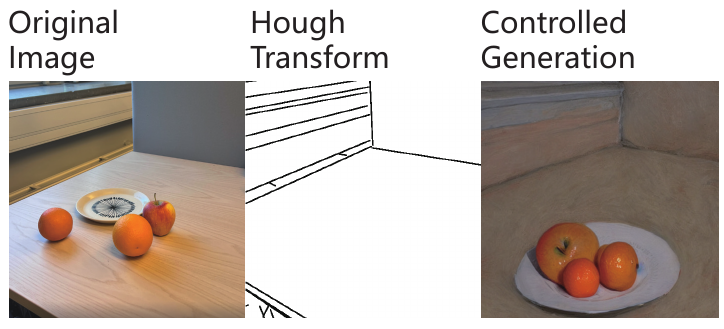}
\caption{Controlled generation with Hough transform.}
\label{fig:controlnet}
\vspace{-1em}
\end{figure}

\paragraph*{\textbf{Departure to 6-DoF}}
One of the main obstacles confining image-goal-based scene rearrangement to 3-DoF settings is the difficulty in controlling the perspective of generated goals to match real-world settings. Consequently, current methods~\cite{kapelyukh2023dall, gkanatsios2023energy} often produce goal images from a top-down view. Inspired by graphical sketching~\cite{enwiki:1188864212}, we utilize a Hough Line Transform~\cite{duda1972use} to extract lines $C_{h}$ from the real scene RGB image $ x^{real} $ (from $X^{real}$) to align the camera's extrinsic matrix between goal and real scene, as detailed in \textit{Eq.}~\ref{eq: hough} and \textit{Eq.}~\ref{eq: controlnet}. The extracted lines $ C_h $ serve as additional perspective control to guide the generation process of Stable Diffusion using ControlNet~\cite{zhang2023adding}, with the training objective in \textit{Eq.}~\ref{eq:control_ldm}. One example is shown in \textit{Figure.}~\ref{fig:controlnet}. 
\begin{equation}
    C_{h} = \textsc{HoughTransform}(x^{real}),
    \label{eq: hough}
\end{equation}
\begin{equation}
   \text{6-DoF:\quad} x^{goal} = \textsc{ControlNet}(C_{h}, \;P,\; s, \; \beta_{cfg}),
    \label{eq: controlnet}
\end{equation}
where $\beta_{cfg} $ is fixed to 7.5 during generation.


\begin{figure*}[!t]
\centering
\includegraphics[height=5.7cm]{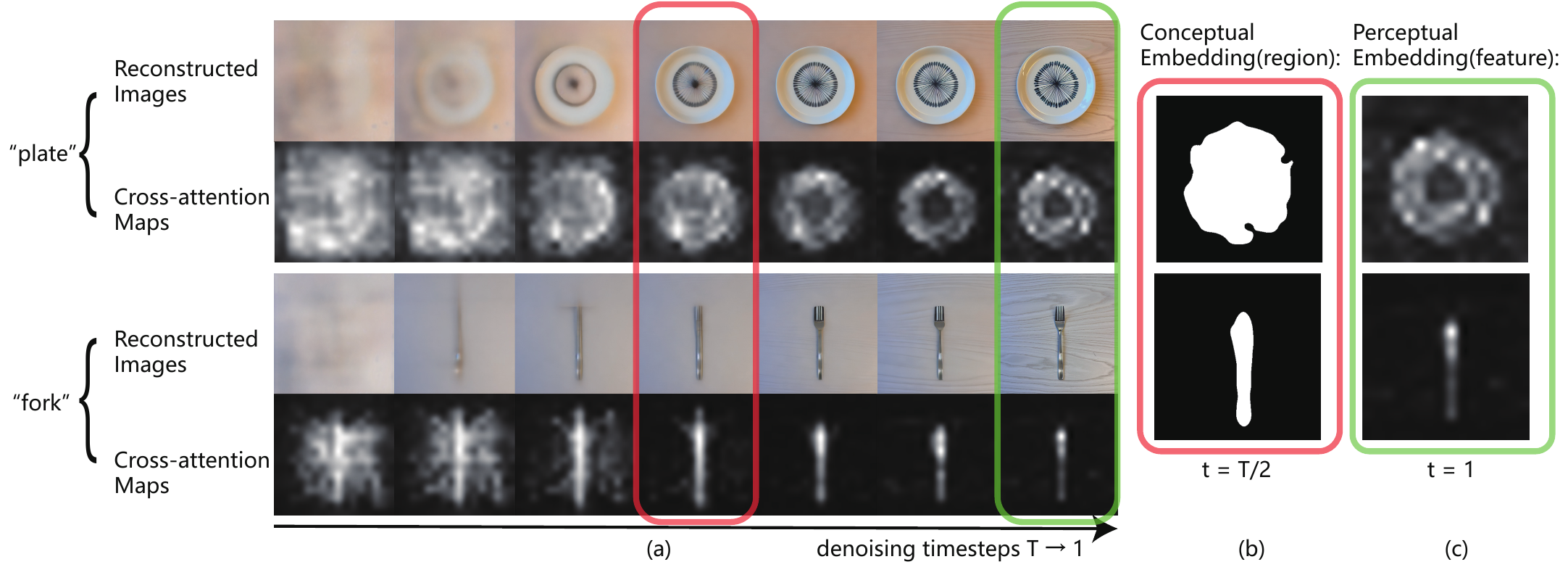}
\caption{(a) Reconstructed images and cross-attention maps \(M(``Plate", t)\), \(M(``Fork", t)\) at different timesteps \(t\) during denoising. (b) Conceptual embedding containing object region information at timestep \(t = T/2\). (c) Perceptual embedding containing object feature information at timestep \(t = 1\). The joint cross-attention representation combines cross-attention from these timesteps using \textit{Eq.}~\ref{joint}. }
\label{fig:observations}
\vspace{-1em}
\end{figure*}

\subsection{Generation with Real Scene $X^{real}$: Thread B}
The representation containing segmentation and feature encoding is produced during the generation. To conduct matching, the representations from the real scene are also required. They are acquired by inverting the real images to their deterministic noise counterparts using DDIM inversion~\cite{song2020denoising}, followed by the standard generation process. 

Formally, we denote by $x^{real}$, a $512 \times 512$ RGB image from the RGBD real scene $X^{real}$, and $z^{real} = \mathcal{E}(x^{real})$ denotes the corresponding encoded latent representation. As shown in \textit{Eq.}~\ref{eq:ddim_inversion}, we invert the clean latent $ z_0^{real} $ into its corresponding noisy counterpart $ z_T^{real} $ through $t : 1 \rightarrow T$ using DDIM inversion~\cite{song2020denoising}, which represents a deterministic special case of DDPM based on ODE limit analysis (\textit{Eq.}~\ref{eq:ddpm_diffusing}):
\begin{equation}
    {z}_t^{real} = \sqrt{\bar{\alpha}_{t}}\frac{{z}_{t-1}^{real} - \sqrt{1 - \alpha_{t-1}}\epsilon_{\theta}}{\sqrt{\bar{\alpha}_{t-1}}} + \sqrt{1 - \alpha_{t}}\epsilon_{\theta}.
    \label{eq:ddim_inversion}
\end{equation}

After inverting the clean latent \( z_0^{real} \) to \( z_T^{real} \) (\textit{Eq.}~\ref{eq:ddim_inversion}), the RGB image captured in the real environment \(x^{real}\) is transformed into its deterministic noise. This allows for the application of \textit{Eq.}~\ref{eq: stablediffusion} for 3-DoF or \textit{Eq.}~\ref{eq: controlnet} for 6-DoF, to reconstruct  \( {z}_0^{real} \) (with \( \beta_{cfg} \) set to 0). During this reconstruction, the zero-shot representation integrating segmentation and feature encoding for the real scene is generated. The reconstructed image can be found in \textit{Figure}~\ref{fig:structure}.

\subsection{Joint Cross-attention Representation}
Apart from representation learning~\cite{rombach2022high}, variational auto-encoders (VAEs)~\cite{kingma2013auto} have also been employed for input data compression~\cite{balle2016end}. \textit{Gregor et al.}~\cite{gregor2016towards} propose a method leveraging VAE for transforming images into a lossy compression with progressively detailed representations. Later, \textit{Ho et al}.~\cite{ho2020denoising} observe that the diffusion model's denoising process resembles progressive autoregressive decoding, with larger image features emerging before finer details. In this paper, we observe that cross-attention maps at different denoising timesteps provide quantification for the above intuition on the lossy nature of diffusion models, as shown in Section~\ref{sec:4.4.1}.


\subsubsection{\textbf{Observations}}
\label{sec:4.4.1}
Based on \textit{Eq.}~\ref{eq:ddpm_arbitrary}, a reconstruction \(\hat{z}_0\) can be estimated at every timestep \( t \) using the additional information received at timesteps \( T \rightarrow t \) during the denoising process:
\begin{equation}
    \hat{z}_0 \approx {z}_0 = \left({z}_t - \sqrt{1 - \bar{\alpha}_t}\epsilon_\theta({z}_t) \right) / \sqrt{\bar{\alpha}_t},
    \label{eq:inverse_arbitrary}
\end{equation} where $ z = \mathcal{E}(x) $.

In the forward process (\textit{Eq.}~\ref{eq:ddpm_diffusing}), noise is gradually added to the clean latent; the noises added during the forward process are predicted and utilized for stepwise denoising (\textit{Eq.}~\ref{ddpm_reverse}). \textit{Viewing the denoising process from an information theory perspective, \textit{Eq.}~\ref{eq:inverse_arbitrary} can be understood as follows: at any time \( t \), partial information \( {z}_t \) becomes available and can be used to estimate the final reconstruction or generation \( \hat{z}_0 \).} In other words, the perceptual detail is gradually added to the generated images with stepwise denoising, similar to the JPEG's progressive encoding~\cite{jpeg2024}. 

We propose using cross-attention maps~\cite{hertz2022prompt} between textual prompts and images to quantify and visualize the intuitions mentioned above. During the training of Stable Diffusion, cross-attention links image-text pairs, reflecting the \textit{correlation between the visual image and textual prompt}. This offers a valuable opportunity to extract knowledge from web-scale models trained with attention mechanisms~\cite{vaswani2017attention} in a zero-shot manner. The cross-attention map $M$ and cross-attention are defined as: 
\begin{equation}
    M = \textsc{Softmax}\left(QK^T/\sqrt{d}\right),
    \label{eq:cross_attn}
\end{equation}
\begin{equation}
    \textsc{Cross-Attention} =MV,
\end{equation}
where \( Q = W^{(i)}_Q \cdot \phi_{z}(z_t) \), \( K = W^{(i)}_K \cdot p \), and \( V = W^{(i)}_V \cdot p \), \(p\) is the encoded user's prompt \(P\). The term \( \phi_{z}(z_t) \) represents the flattened \( z_t \), and \( W^{(i)}_Q \), \( W^{(i)}_K \), and \( W^{(i)}_V \) are the projection matrices. \(M\) here can be indexed by the object's word \(w \in P\) and timestep \(t \in T\), for example, \(M(w, t)\) refers to cross-attention maps of the object's word \(w\) at timestep \(t\) during denoising \( T \rightarrow 1 \), with a dimension of \(512 \times 512\). Cross-attention maps show the similarity between $Q$ and $K$.

The reconstruction and cross-attention maps with respect to each word at various timesteps during the DDIM denoising process can be found in \textit{Figure}~\ref{fig:observations}.  \textit{We observe that during the denoising process, the object's shape (or region) typically becomes apparent first in the reconstructed images, followed by the emergence of texture (or feature) details. In the corresponding cross-attention maps, the activations appear to quantify this trend.} For instance, in the cross-attention maps for a ``fork", the activation initially highlights the general shape and later becomes concentrated around the fork's head, where features are most distinct.

\subsubsection{\textbf{Representation construction}}
Building on the observed characteristics of the lossy denoising process, we propose a zero-shot joint cross-attention representation. Following the terminology in \cite{gregor2016towards}, we define two stages that preserve different attributes of an object:

\begin{itemize}
    \item Conceptual Embedding (region): Captures the object's conceptual region, useful for segmentation, obtained at the mid-stage stage of DDIM denoising when \( t \approx T/2 \).
    \item Perceptual Embedding (feature): Captures the object's feature-rich region, useful for matching, obtained towards the end of the DDIM denoising when \( t \approx 1 \).
\end{itemize}

We define the joint cross-attention representation, combining conceptual and perceptual embeddings, as:
\begin{equation}
    R(w) = \textsc{Step}(M(w,\; T/2), \tau_1) + \textsc{Step}(M(w,\; 1), \tau_2)
    \label{joint}
\end{equation}
Here, \(\textsc{Step}\) is a Heaviside step function with thresholds \(\tau_1 = 0.3\) and \(\tau_2 = 0.9\), designed to filter out random noise and to retain only feature-rich areas, respectively, thus reducing computational complexity during the matching process. The activation of \(M(w)\) ranges between 0 and 1, \(T\) is fixed to 50. Together, these form an object-level representation integrating generation, segmentation and feature encoding. Examples can be found in \textit{Figure.}~\ref{fig:observations} (b) and (c).

\subsection{Object Matching and Robot Execution}
An optimal transformation \( T_{\text{opt}}(w) \) is determined for each object \( w \) by minimizing the squared difference between the goal and real representations:
\begin{equation}
    T_{\text{opt}}(w) = \underset{T}{\text{argmin}}\left\| R^{\text{Goal}}(w) - T(R^{\text{Real}}(w)) \right\|^2,
\end{equation}
where the matching process is conducted within each object's representation \( R(w) \). The Iterative Closest Point (ICP) algorithm~\cite{besl1992method} and Marigold algorithm~\cite{ke2023repurposing} are employed to determine the object's relative pose transformation \( T = (x,\;y,\;z, \;\theta) \), with the initial pose being calculated from \( R^{\text{Real}}(w) \). Implementation details are in Section~\ref{sec:5.6}.

\section{Experiments}
\subsection{Experimental Evaluation}
Given the highly modular nature of the scene rearrangement task, we analyze goal generation, matching and robot execution separately with the following questions:
\begin{enumerate}
    \item Can web-scale generative models produce goals that are satisfactory to humans?
    \item Do categorical descriptions yield better generations than specific one-to-one correspondences in terms of distortion and image-prompt alignment?
    \item Does a joint representation integrating generation, segmentation and feature encoding have a higher matching accuracy than the multi-step design?
    \item Can \textit{PACA} perform scene rearrangement in a zero-shot manner?
\end{enumerate}
\subsection{Experiment Setup}
\subsubsection{Robotic Platform}
\begin{figure*}[!t]
\centering
\includegraphics[height=5.5cm]{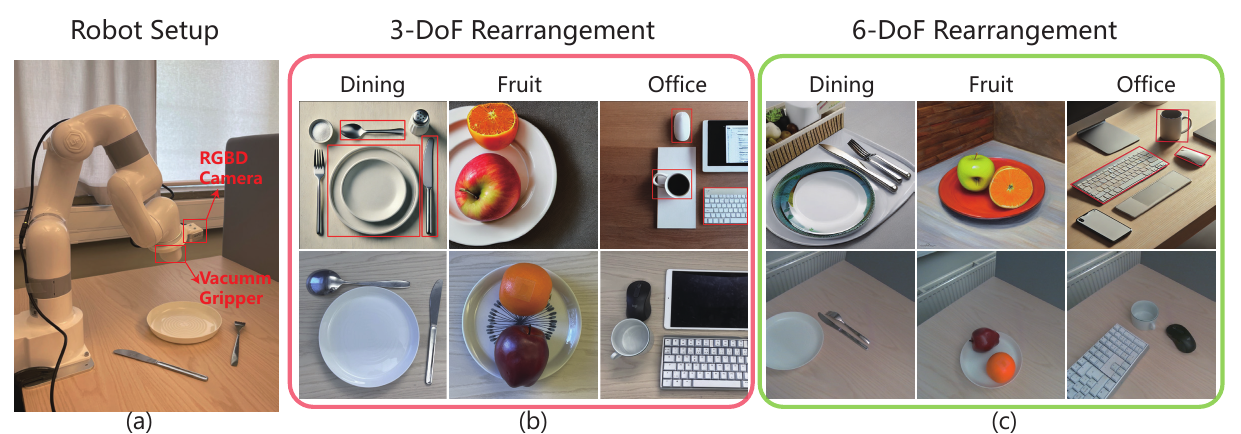}
\caption{(a) Experimental setup; (b) Examples of 3-DoF top-down rearrangement; (c) Examples of 6-DoF rearrangement. In (b) and (c), the first row is the goal, the second row shows the final rearrangement. Objects relevant to the prompt are marked with red boundaries due to stochasticity, distortions, and new object generation. Objects not specified in the prompt do not affect the pipeline, as they lack object-level representation. Full executions can be found in the supplementary material. }
\label{fig:experiment}
\vspace{-1em}
\end{figure*}

We deploy \textit{PACA} on a 6-DoF UFactory Lite 6 robot arm equipped with a vacuum gripper and a wrist-mounted Intel RealSense D405 RGBD camera to conduct 3-DoF and 6-DoF rearrangements, as shown in \textit{Figure}~\ref{fig:experiment} (a). For grasping, we position the robot arm with a wrist-mounted RGBD camera above the table, select the geometric center of the object's representation as the grasping point, and perform top-down grasping using a vacuum gripper.
\subsubsection{Tasks}
\label{sec:5.2.2}
Three everyday tabletop rearrangement tasks are adopted from \cite{kapelyukh2023dall}: 1) Dining Scene: This scene features randomly placed knives, forks, spoons, and one plate; 2) Fruit Scene: This scene contains randomly placed apples, oranges, and one plate; 3) Office Scene: This scene includes randomly placed keyboards, mice, and one mug.

\subsubsection{Baselines}
\begin{itemize}
    \item SG-Bot~\cite{zhai2023sg} generates goals from scene graphs by training Graph-to-3D~\cite{dhamo2021graph} and AtlasNet~\cite{groueix2018papier}. We use their trained models for the experiments in Section~\ref{sec: 5.3}.
    
    \item DALL·E-Bot~\cite{kapelyukh2023dall}, which is the work most closely related to ours, uses DALL·E 2~\cite{ramesh2022hierarchical} to generate goal images, with subsequent matching via Mask R-CNN~\cite{he2017mask} segmentation and CLIP~\cite{radford2021learning} feature encoding. We replicate their approach using the same prompts for experiments in Sections~\ref{sec:5.4} and ~\ref{sec:5.5}.
\end{itemize}

\subsection{Are goals generated by web-scale models satisfactory to humans?}
\label{sec: 5.3}
A user study is conducted to evaluate user satisfaction with rearrangement goals from two humans, \textit{PACA}, SG-Bot, and a custom baseline \textit{Parallel}, where objects were horizontally aligned. The study include 3 scenes from Section~\ref{sec:5.2.2} with 43 subjects providing 559 ratings on a scale of 1 (Very bad) to 5 (Very good), as shown in \textit{Table.}~\ref{user}. The results show that \textit{PACA} achieves user satisfaction comparable to the two human users, whose scores vary likely due to individual preferences. SG-Bot potentially struggles with overlap between objects and asymmetry in the generated goal. Due to the lack of trained models and simulated objects, Fruit and Office scenes are inapplicable to SG-Bot. Implementation details and the full user study are available in the supplementary material.

\begin{table}[h]
\centering
\setlength{\tabcolsep}{2pt}
\scalebox{0.9}{
\begin{tabular}{lccccc}
\toprule
&$\text{Human 1}$ & $\text{Human 2}$ & \textit{PACA} & \textit{Parallel} & SG-Bot \\
\midrule
Dining & {3.41} & {\textbf{4.34}} & {\underline{3.85}} & {3.05} & {1.68} \\
Fruit & {\textbf{4.63}} & {1.89} & {\underline{4.42}} & {2.26} & --- \\
Office & {\underline{3.55}} & {\textbf{3.67}} & {3.49} & {1.95} & --- \\
\bottomrule 
\end{tabular}
}
\caption{Average ratings for \textit{PACA} and four baselines, with the highest displayed in bold and the second highest underscored.} 
\label{user}
\end{table}

\subsection{Are categorical descriptions more effective?}
\label{sec:5.4}
In this section, we examine whether prompts referring to object categories (e.g. apple) can lead to better goal generation compared to specific one-to-one correspondence (e.g. an apple) prompts in DALL·E-Bot, using the following metrics:
\begin{itemize}
\item Inception Score (IS)~\cite{salimans2016improved}: Evaluates the quality of images generated by generative adversarial networks (GANs), focusing on distortion levels.
\item CLIP Score (CLIP)~\cite{hessel2021clipscore}: Measures the similarity between a generated image and its textual caption, assessing image-prompt alignment and relevance.
\end{itemize}

The results, detailed in \textit{Table.}~\ref{table:prompt}, are based on tests conducted on 2,000 images generated with Stable Diffusion across each scene. The results indicate that categorical prompts yield generation with less distortion and better alignment with the prompts compared to specific descriptions. The difference in scores between scenes is likely due to the varying sizes of training datasets across the scenes.



It can be observed that \textit{using categorical descriptions of the scene's contents yields better generations than specific one-to-one correspondence descriptions, in terms of distortion and image-prompt alignment. This approach may reduce the need for human intervention in goal selection}. This is likely influenced by the image-caption pairs used to train these generative models; for example, Stable Diffusion was trained on the LAION-5B dataset\cite{schuhmann2021laion}, which predominantly utilizes categorical descriptions as captions.

\begin{table}[!h]
\centering
\setlength{\tabcolsep}{1pt}
\scalebox{0.75}{
\begin{tabular}{lcccc}
\toprule
\textbf{Scene} & \textbf{Prompt} & \textbf{IS\(\uparrow\)} & \textbf{CLIP\(\uparrow\)}\\
\midrule
\multirow{2}{*}{Dining}  & fork, knife, spoon, plate, table & \bm{$4.08 \pm 0.21$} & \textbf{31.19}\\
 & A fork, a knife, a plate, and a spoon, top-down & {$3.54 \pm 0.24$} & 30.33 \\
\midrule
\multirow{2}{*}{Fruit} & apple, orange, plate & \bm{$3.53 \pm 0.19$} & \textbf{21.54}\\
  & Two apples, and an orange, top-down & {$2.99 \pm 0.15$} & 19.80 \\
 \midrule
 \multirow{2}{*}{Office} & keyboard, mouse, mug & \bm{$3.81 \pm 0.20$} & \textbf{20.54}\\
 & A keyboard, a mouse, and a mug, top-down &{$3.55 \pm 0.21$} & 20.32\\
\bottomrule
\end{tabular}
}
\caption{Quantitative evaluation of prompt types over 6,000 generated images. The first and the second rows in each scene are categorical description and specific description, respectively. }
\label{table:prompt}
\end{table}

\subsection{Does joint representation perform better than multi-step design?}
\label{sec:5.5}
In this section, we compare the joint representation and modular design across three scenes and assess the matching performance. The real images are taken under natural illumination and feature objects with varying textures. These are to be matched with scenes generated using Stable Diffusion, containing the same objects as described in the categorical descriptions above. Each scene includes 150 matching pairs. As performance metric, we consider:
\[
\text{Matching\; Accuracy} = \sum_{i=1}^{150} \frac{N_{\text{acc}}^i}{N_{\text{total}}^i},
\]
where \(N_{\text{total}}^i\) represents the total number of matchings in a real-goal pair, and \(N_{\text{acc}}^i\) is the number of successfully matched objects. 
The matching results are shown in \textit{Table}~\ref{table: matching}. Our integrated design outperforms the multi-step design to a noticeable extent. This improvement is largely due to distortions that commonly occur in generated images. These distortions result in repetitive and inaccurate croppings in the initial segmentation \cite{he2017mask}, and errors are then magnified during the image encoding process \cite{radford2021learning}, which degrades performance at each step. Meanwhile, the joint representation, which integrates generation, segmentation, and feature encoding, reduces this multi-step error accumulation and maintains more consistent performance with both real images and generated outputs.

\begin{table}[h]
\centering
\setlength{\tabcolsep}{2pt}
\scalebox{0.9}{
\begin{tabular}{lcccc}
\toprule
Method & Dining & Fruit & Office & Average \\ 
\midrule
DALL·E-Bot & 42.83\% & 42.57\% & 68.66\% & 51.35\% \\ 
Ours & 83.75\% & 88.63\% & 87.78\% & 86.72\% \\
\bottomrule 
\end{tabular}
}
\caption{Quantitative evaluation of matching accuracy over 450 pairs of images.}
\label{table: matching}
\end{table}

\subsection{Can \textit{PACA} be performed in a zero-shot manner?}
\label{sec:5.6}
We evaluate the zero-shot performance of \textit{PACA} in 3-DoF and 6-DoF settings. In 3-DoF, success rates are derived from a comparison of generated goals and robot execution images based on absolute positions and rotations of the objects. Each experiment is conducted with 150 pairs per scene. In 6-DoF, object depth is first estimated with Marigold~\cite{ke2023repurposing} in the generated goal and then aligned with actual depth from an RGBD camera by minimizing the following L2 loss:
\begin{equation}
    L(\theta) = \|D_{est}(\theta) - D_{real}\|^2,
\end{equation}
where $D_{est}(\theta)$ represents the depth estimated by Marigold, and $D_{real}$ is the depth measured by the RGBD camera, and $\theta$ stands for the scaling and shifting parameters. We conducted ten random object position initializations in each scene, resulting in a total of 60 trials. Due to the absence of ground truth, the criterion for a successful rearrangement is that the relative positions between the objects must align with those of the generated goal, although slight variances in absolute positions are permissible. The results can be found in \textit{Table}~\ref{table: execution} and \textit{Figure}~\ref{fig:experiment} (b) and (c). 
The lower execution success rate in the 6-DoF setting compared to the 3-DoF setting is likely due to errors in the estimated depth of the goal's objects, which could lead to inaccurate position during grasping and placing.
\begin{table}[h]
\centering
\setlength{\tabcolsep}{2pt}
\scalebox{0.9}{
\begin{tabular}{lcccc}
\toprule
Setting & Dining & Fruit & Office & Average\\
\midrule
3-DoF & 61.93\% & 64.61\% &79.67\% & 68.73\%\\
6-DoF & 56.67\% & 66.67\% &73.33\% & 65.56\%\\
\bottomrule 
\end{tabular}
}
\caption{Success rates of \textit{PACA} in 3-DoF and 6-DoF settings.}
\label{table: execution}
\vspace{-1em}
\end{table}

\subsection{Discussion: Are there other potential use cases?}
\begin{figure}[h]
\centering
\includegraphics[height=6.4cm]{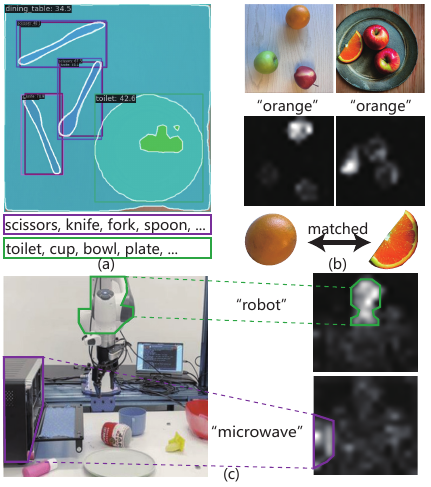}
\caption{(a) Results from Mask R-CNN~\cite{he2017mask}, with text indicating the predicted labels. (b) An example of cross-domain matching showcasing zero-shot performance. (c) An example of the joint cross-attention representation applied in a robotic environment from CACTI~\cite{mandi2022cacti}, where the representation can be used for masking or image editing for data augmentation.}
\label{fig:others}
\end{figure}

Before exploring additional potential use cases, we first summarize the key features of our joint cross-attention representation:
\begin{itemize}
    \item \textbf{Alleviation of multi-step error accumulation}, as discusses in Section~\ref{sec:5.5}. 
    \item \textbf{Effective bridge} between the real scene and the generated goal. Since the Stable Diffusion is trained with massive real-world data, our representation leveraging Stable Diffusion is effective in capturing the feature in both the real scenes and the generated goals. 
    \item \textbf{Strong contextual information}. The usage of the categorical prompt provides a strong context for feature encoding, which can largely enhance accuracy in matching contexts (\textit{Figure.}~\ref{fig:others} (a)).
    \item \textbf{Cross-domain matching}. Compared to classical matching methods such as color histograms \cite{swain1991color}, keypoints \cite{lowe1999object}, and templates \cite{brunelli2009template}, \textit{PACA} can realize cross-domain matching when the object belongs to the same concept (\textit{Figure.}~\ref{fig:others} (b)).
\end{itemize}
Based on the above traits, this representation has potential in the field of robotic manipulation using text-driven generative models. For example, when learning from synthetic videos~\cite{ko2023learning, black2023zero}, this representation may be helpful in matching the robotic arm in the real scene with the generated goal, facilitating action derivation.  Other potential use cases include data augmentation~\cite{mandi2022cacti, zhang2024diffusion, chen2024mirage, yu2023scaling}, where different objects need to be segmented out before conducting background transformations using generative models. One example can be found in \textit{Figure.}~\ref{fig:others} (c).
\vspace{-1em}

\section{Conclusions and Limitations}
\paragraph{Limitations}
We provide a full pipeline for image-goal-based scene rearrangement; however, two aspects of the pipeline, in particular, require further improvement: 1) While categorical descriptions enhance generation quality, they compromise on user personalization by not fully capturing precise spatial relationships, and the inherent limitations of current text-driven generative models restrict the control over the number of generated objects; 2) In the 6-DoF setting, the ICP~\cite{besl1992method} and Marigold~\cite{ke2023repurposing} algorithms are used to estimate the depth of objects in the generated goal, yet this method does not provide all estimates for critical information needed for effective grasping, such as the center of mass of an object.


\paragraph{Conclusions}
In this paper, we introduce \textit{PACA}, a zero-shot 6-DoF scene rearrangement pipeline, facilitated by a joint cross-attention representation derived from the lossy denoising sampling of diffusion models. Experimental results demonstrate that this representation effectively reduces multi-step error accumulation and enhances matching accuracy, including in the case of cross-domain samples. Additionally, perspective control within this framework supports 6-DoF operations using image-based goals. Future work will focus on utilizing energy-based models to refine control over image generation, and precisely matching the prompts in terms of number and spatial relationships, thereby enhancing personalization capabilities. Another direction is to integrate depth estimation within the representation construction to enable more accurate object grasping and placement.
{\small
\balance
\bibliographystyle{ieee_fullname}
\bibliography{citations}
}

\end{document}


\title{Supplementary Material\\PACA: \underline{P}erspective-\underline{A}ware \underline{C}ross-\underline{A}ttention Representation \\ for Zero-Shot Scene Rearrangement}  

\author{
Shutong Jin$^{1*}$, Ruiyu Wang$^{1*}$, Kuangyi Chen$^2$, Florian T. Pokorny$^1$ \\
$^1$KTH Royal Institute of Technology, \ $^2$Graz University of Technology\\
{\tt\small \{shutong, ruiyuw, fpokorny\}@kth.se, kuangyi.chen@tugraz.at}\\
}

\maketitle
\thispagestyle{empty}
\appendix
\section{User Study}
\subsection{Overview}
A user study was conducted to evaluate satisfaction with rearrangement goals generated by two humans, \textit{PACA}, SG-Bot~\cite{zhai2023sg}, and a custom baseline called \textit{Parallel}. The study involved 43 subjects, who rated 3 scenes from Section \textit{5.2.2}, providing a total of 559 ratings on a scale from 1 (Very bad) to 5 (Very good). During the study, subjects were asked, ``If a robot made this rearrangement for you at home, how satisfied would you be?" The full results, including mean and standard deviations, are presented in \textit{Table}~\ref{user}.  The following sections outline the implementation of the user study and the baseline methods.

\subsection{Implementation of the User Study}
\begin{figure}[h]
\centering
\includegraphics[width=0.56\linewidth]{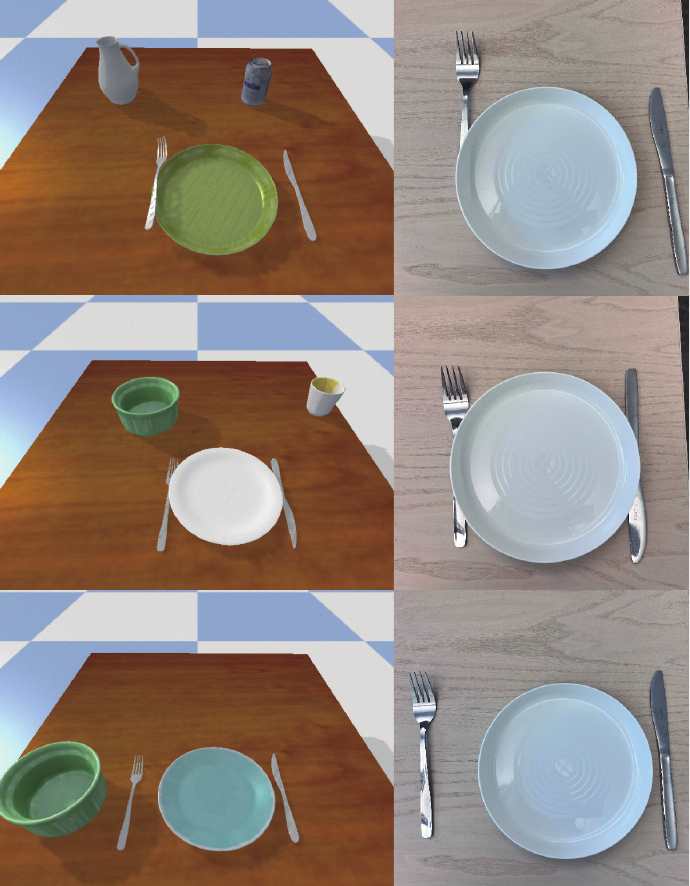}
\caption{Scene Dining of SG-Bot: The first column shows the generated goals and the second column shows the human rearrangements based on those goals.}
\label{sgbot}
\end{figure}

Given the highly modular nature of the scene rearrangement task, this user study focuses solely on analyzing user satisfaction with the generated goals, independent of the matching and robot execution stages. To minimize the influence of users' ratings based on the visual appearance of the goals, we recreated the object arrangements in the real world to match their positions in the simulated or synthetic images and presented only the unified real-world images to the users. For example, SG-Bot’s goals were originally generated in simulation. The corresponding recreated scenes are shown in \textit{Fig.}~\ref{sgbot}.

\begin{figure}[h]
\centering
\includegraphics[width=1.01\linewidth]{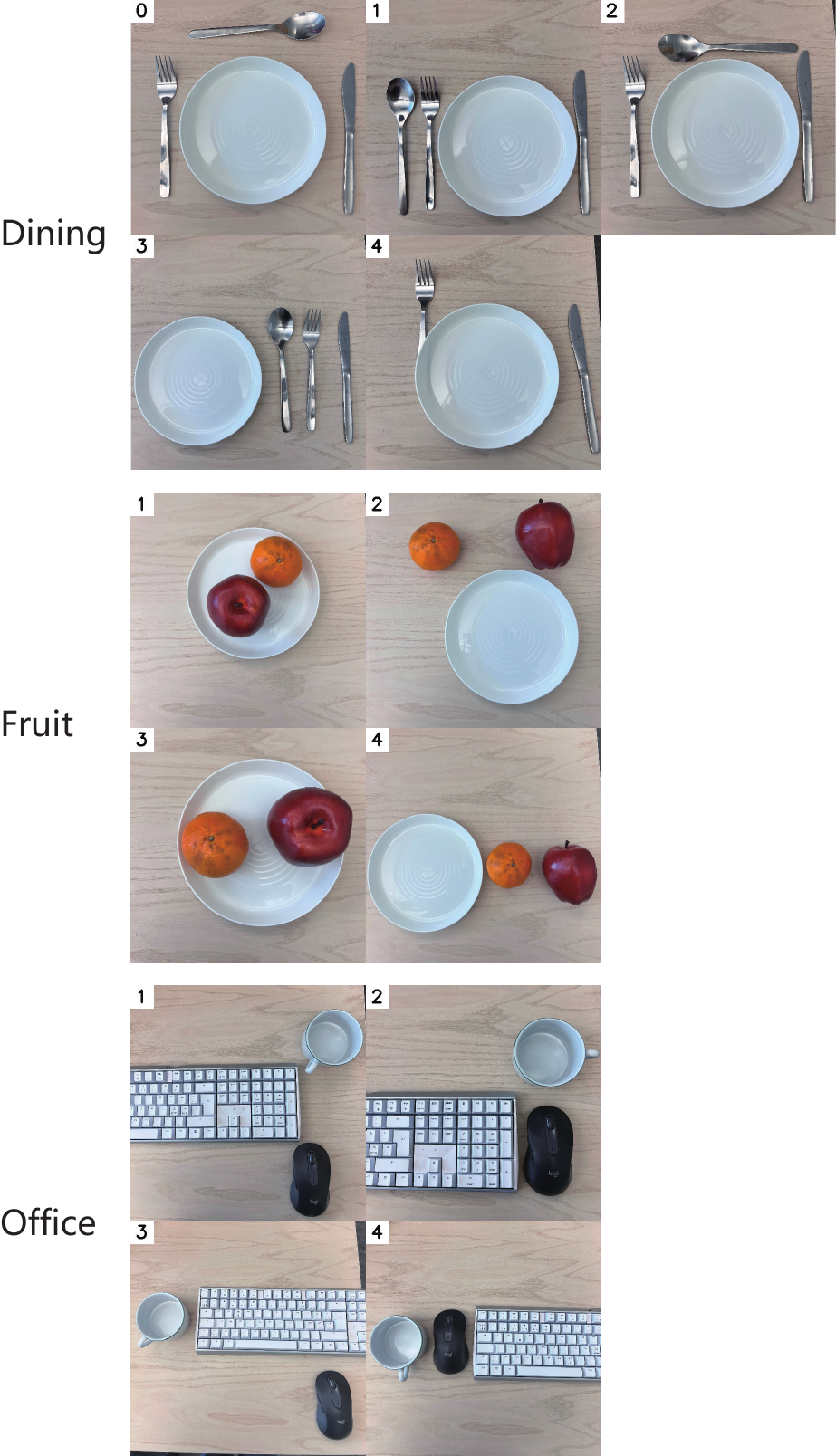}
\caption{An example of a user study sent to a subject.}
\label{userstudy_example}
\end{figure}

For each subject, the study randomly sampled one goal from each baseline for each scene and anonymously numbered the rearrangements. The subject was then asked to rate the numbered images. An example of the user study sent to one subject is shown in \textit{Fig.}~\ref{userstudy_example}.
Due to the lack of trained models and simulated objects, Fruit and Office scenes are inapplicable to SG-Bot.

\subsection{Implementation of the Baselines}
\subsubsection{$\text{Human 1}$ and $\text{Human 2}$}
\begin{figure}[h]
\centering
\includegraphics[width=0.8\linewidth]{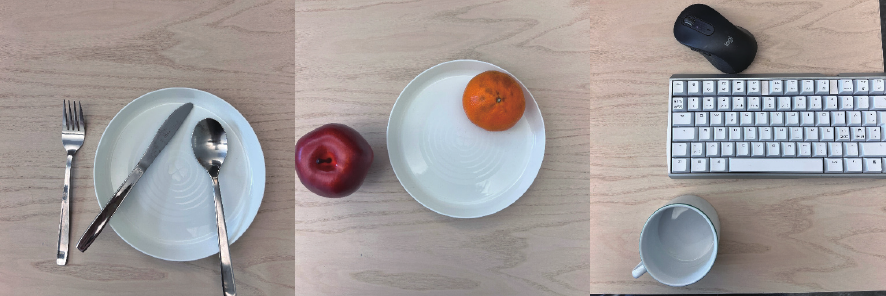}
\caption{Initial object states for three scenes. }
\label{initialization}
\end{figure}

\begin{figure}[h]
\centering
\includegraphics[width=0.56\linewidth]{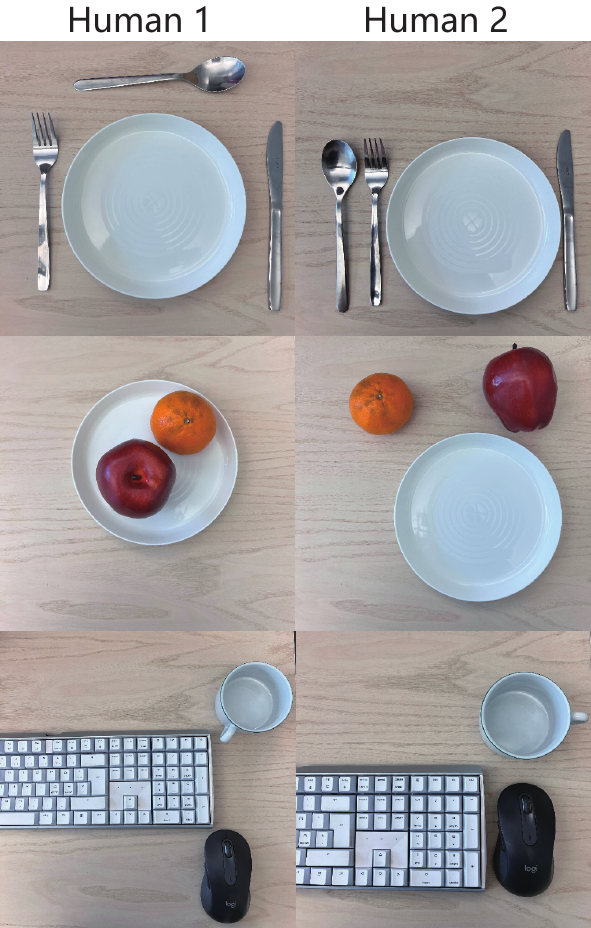}
\caption{Rearrangements of three scenes performed by two humans.}
\label{humans}
\end{figure}

\textit{Fig.}~\ref{initialization} shows the initial setup of the three scenes, with objects randomly scattered on the table. Before each human rearrangement, the scene was reset to this state. The best-effort arrangements were carried out by the first two authors (Human 1 and Human 2), each working independently without seeing the other's results. The outcomes are shown in \textit{Fig.}~\ref{humans}.

\subsubsection{\textit{PACA}}
\begin{figure}[h]
\centering
\includegraphics[width=0.56\linewidth]{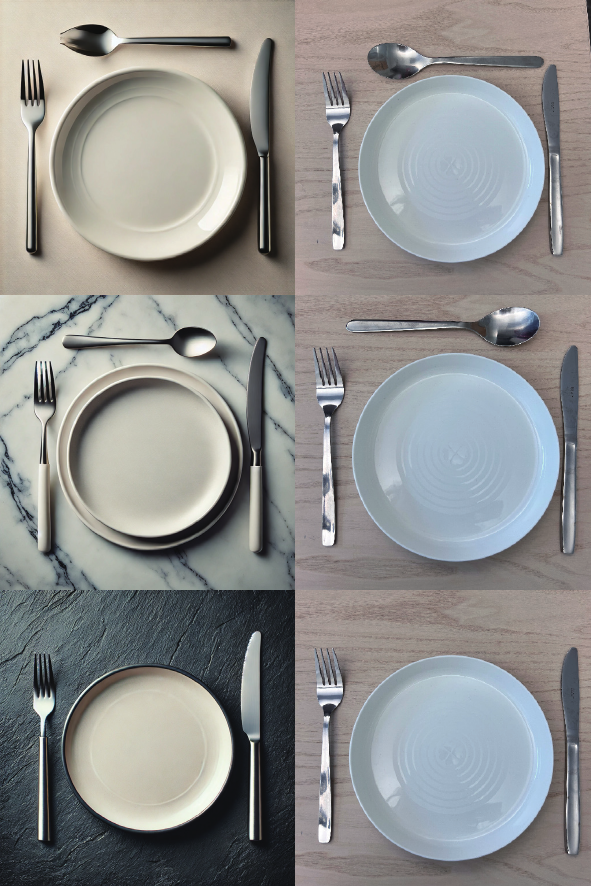}
\caption{Scene Dining of \textit{PACA}: The first column shows the generated goals and the second column shows the human rearrangements based on those goals.}
\label{paca_dining}
\end{figure}

\begin{figure}[h]
\centering
\includegraphics[width=0.56\linewidth]{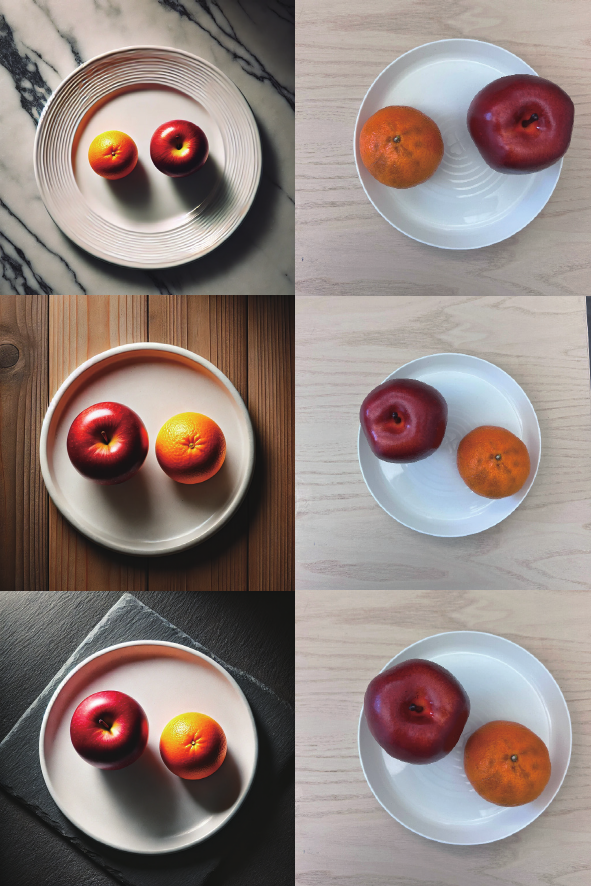}
\caption{Scene Fruit of \textit{PACA}: The first column shows the generated goals and the second column shows the human rearrangements based on those goals.}
\label{paca_fruit}
\end{figure}

\begin{figure}[h]
\centering
\includegraphics[width=0.56\linewidth]{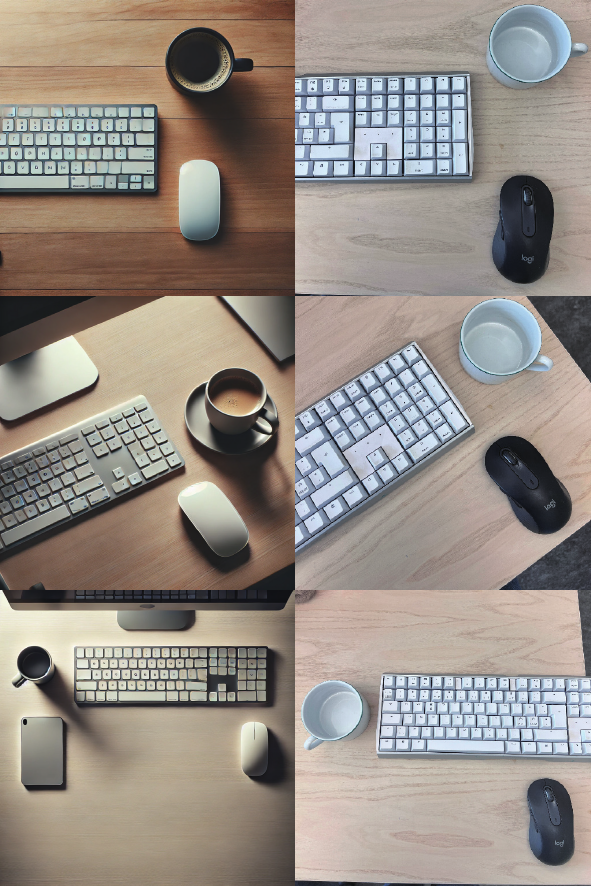}
\caption{Scene Office of \textit{PACA}: The first column shows the generated goals and the second column shows the human rearrangements based on those goals.}
\label{paca_office}
\end{figure}

Three goals for each scene were generated using the prompts listed in Section \textit{5.4}. An additional set was created with the prompt ``fork, knife, plate, table" (excluding ``spoon") to align with the goals generated by SG-Bot, where the spoon is absent. The goals for each scene are shown in \textit{Fig.}~\ref{paca_dining}, \textit{Fig.}~\ref{paca_fruit}, and \textit{Fig.}~\ref{paca_office}, respectively.

\subsubsection{\textit{Parallel}}
\begin{figure}[h]
\centering
\includegraphics[width=0.8\linewidth]{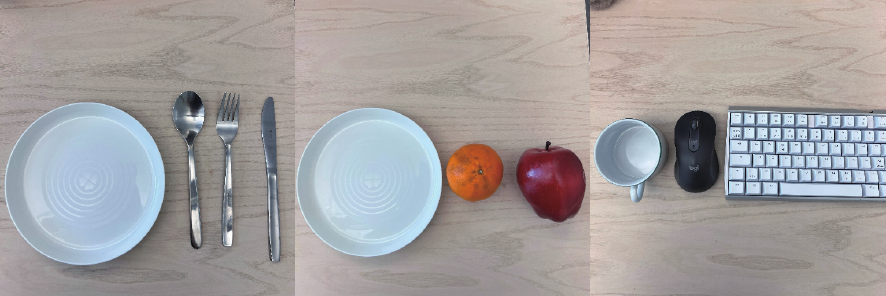}
\caption{Rearrangements of three scenes in \textit{Parallel}.}
\label{parallel}
\end{figure}

\textit{Parallel} is a custom rearrangement baseline in which objects were manually aligned horizontally in random order without overlapping, as shown in \textit{Fig.}~\ref{parallel}.
\subsubsection{SG-Bot}
SG-Bot generates goals from scene graphs by training Graph-to-3D~\cite{dhamo2021graph} and AtlasNet~\cite{groueix2018papier}. Trained in the PyBullet~\cite{coumans2016pybullet} environment, SG-Bot is not a zero-shot method and focuses specifically on Dining scenes, with no trained models or simulated objects for Fruit and Office scenes. Therefore, we only compared their results for the Dining scene. The examples used in the user study are shown in \textit{Fig.}~\ref{sgbot}. We obtained the goals by running their open-source code for evaluation and demonstration without modifying model weights or inputs. For the user study, we selected the demonstrations that are most closely aligned with our scenes.

\subsection{Table}
See page 2.
\FloatBarrier
\begin{table*}[!h]
\centering
\setlength{\tabcolsep}{2pt}
\scalebox{1}{
\begin{tabular}{lccccc}
\toprule
&$\text{Human 1}$ & $\text{Human 2}$ & \textit{PACA} & \textit{Parallel} & \text{SG-Bot} \\
\midrule
Dining & {$3.41 \pm 1.23$} & \bm{$4.34 \pm 0.81$} & \underline{$3.85 \pm 1.28$} & {$3.05 \pm 0.85$} & {$1.68 \pm 0.90$} \\
Fruit & \bm{$4.63 \pm 0.67$} & {$1.89 \pm 1.12$} & \underline{$4.42 \pm 0.82$} & {$2.26 \pm 1.25$} & --- \\
Office & \underline{$3.55 \pm 1.32$} & \bm{$3.67 \pm 1.27$} & {$3.49 \pm 1.23$} & {$1.95 \pm 0.97$} & --- \\
\bottomrule 
\end{tabular}
}
\caption{Mean and standard deviations of the ratings for \textit{PACA} and four baselines, with the highest values in bold and the second highest underlined.} 
\label{user}
\end{table*}

\subsection{Figures}
See page 2 and 3.
{\small
\bibliographystyle{ieee_fullname}
\bibliography{citations}
}